\documentclass[preprint,3p]{elsarticle}

\usepackage{lineno,hyperref}
\usepackage{xcolor}
\modulolinenumbers[5]

\begin{document}

\begin{frontmatter}

\title{Correlated-informed neural networks: a new machine learning framework to predict pressure drop in micro-channels}


\author[mymainaddress]{J.A. Montañez-Barrera\corref{mycorrespondingauthor}}
\cortext[mycorrespondingauthor]{Corresponding author}
\ead{ja.montanezbarrera@ugto.mx}
\author[address2]{J.M. Barroso-Maldonado}
\author[address3]{A.F. Bedoya-Santacruz}
\author[address4]{Adrián Mota-Babiloni}

\address[address1]{Engineering Division, Irapuato-Salamanca campus, University of Guanajuato, Salamanca, Gto 36885, Mexico}
\address[address2]{CETYS University, Engineering College, Mexicali, BC, 21259, Mexico.}
\address[address3]{Department of Electromechanical engineering, University Pedagogic and Technologic of Colombia, Duitama, Boyacá 150461, Colombia.}
\address[address4]{ISTENER Research Group, Department of Mechanical Engineering and Construction, Universitat Jaume I (UJI), Castelló de la Plana E-12071, Spain}
\begin{abstract}
Accurate pressure drop estimation in forced boiling phenomena is important during the thermal analysis and the geometric design of cryogenic heat exchangers. However, current methods to predict the pressure drop have one of two problems: lack of accuracy or generalization to different situations. In this work, we present the correlated-informed neural networks (CoINN), a new paradigm in applying the artificial neural network (ANN) technique combined with a successful pressure drop correlation as a mapping tool to predict the pressure drop of zeotropic mixtures in micro-channels. The proposed approach is inspired by Transfer Learning, which is highly used in deep learning problems with reduced datasets. Our method improves the ANN performance by transferring the knowledge of the Sun \& Mishima correlation for the pressure drop to the ANN. The correlation having physical and phenomenological implications for the pressure drop in micro-channels considerably improves the performance and generalization capabilities of the ANN. The final architecture consists of three inputs: the mixture vapor quality, the micro-channel inner diameter, and the available pressure drop correlation. The results show the benefits gained using the correlated-informed approach predicting experimental data used for training and a posterior test with a mean relative error ($mre$) of 6\%, lower than the Sun \& Mishima correlation of 13\%. Additionally, this approach can be extended to other mixtures and experimental settings, a missing feature in other approaches for mapping correlations using ANNs for heat transfer applications.
\end{abstract}

\begin{keyword}
two-phase flow \sep pressure drop \sep zeotropic mixtures \sep machine learning \sep transfer learning \sep ANN \sep micro-channels.
\end{keyword}

\end{frontmatter}


\setlength{\tabcolsep}{18pt}

\begin{table*}[h]
\centering
\begin{tabular}{llll}
\multicolumn{2}{l}{\textbf{Nomenclature}} & \multicolumn{2}{l}{\textbf{Greek Symbols}}\\
{\it ID} & Internal Diameter (m) & $\alpha$ & ANN input vector \\
$D_h$ & Hydraulic Diameter (m) & $\delta$ & ANN output\\
$G$ & Mass flux (kg/s-m$^2$) & $\xi$ & Roughness ($\mu$m) \\
$g$ & Gravity (m/s$^2$)& $\mu$ & Dynamic viscosity (Pa-s) \\
$L$ & Length (m)& $\rho$ & Density (kg/m$^3$) \\
$La$ & Laplace number & $\sigma$ & Surface tension \\
$f$ & Friction factor &  \\
$mre$ & Mean relative error & \\
$mse$ & Mean squared error &  \\
ANN & Artificial Neural Network & \\
PCC & Pearson Correlation Coefficient & \\
AAD & Absolute Average Deviation & \\
$N$ & Number of samples &\multicolumn{2}{l}{\textbf{Subscripts}} \\
$P$ & Pressure (kPa) & {\it eff} & Effective \\
$Re$ & Reynolds number & $K$ & Target neuron \\
$T$ & Temperature (K)& $l$ & Liquid phase \\
$X$ & Martinelli Parameter & $mix$ & Mixture\\
$x$ & Vapor quality & $2ph$ & Two-phase \\
$w$ & Weights of the ANN & $v$ & Vapor phase \\
\end{tabular}
\end{table*}

\section{Introduction}

Throughout the history of Mechanical Engineering, modeling physical phenomena have been a challenge for the scientific community, which remains unchanged nowadays. Physical and phenomenological modeling can be differenced mainly by its results’ quality. Meanwhile, physical modeling offers a generalization of a phenomenon. It commonly deviates from experiments. On the other hand, phenomenological modeling accurately describes specific phenomena but lacks generalization. Despite such differences, both are extensively used by researchers.

Phenomenological models trade-off their simplicity features given by their easy access and programming, with its limitation of covering a wide range of scenarios (the opposite features are presented by physical modeling). In transport phenomena and fluid mechanics, the phenomenological models are the most used to compute different parameters. For instance, heat transfer coefficients, pressure drop, isentropic efficiencies, volumetric efficiencies, losses, etc., are essential parameters in designing, analyzing, and simulating mechanical devices and their systems. In the case of pressure drop prediction applied in low-temperature fields, the relative error of the available methods averages 20\% \cite{Awad2008, Sun2009, Cicchitti1959}. Hence, the trend of analyzing such models prioritizes two research lines, improving the existing methods or correlations, and introducing new correlations based on experimental data. 

For instance, Asadi et al. \cite{Asadi2014} reviewed pressure drops correlations for two-phase flow in micro-channels, and Barraza et al. \cite{Barraza2016} used some of these correlations to verify their applicability in experimental models of cryogenic boiling of non-azeotropic mixtures. These experiments were performed in mini channels while varying the mass flux, the roughness, the pressure, the tube diameter, and the mixture composition. From their results, three correlations describe the pressure drop with the lowest error, the Awad and Muzynchka \cite{Awad2008}, the Sun and Mishima \cite{Sun2009}, and Chiccitti et al. \cite{Cicchitti1959}. Moharana et al. \cite{Moharana2013} used a two-dimensional semi-analytical model to estimate the pressure drop in singly- and doubly connected ducts.

In the proposal of new methods, in the last decade (2010-2020), researchers have proposed alternative phenomenological models based on machine learning to assist the mechanical engineering in complex phenomena related to fluid dynamics, being the Artificial Neural Networks (ANN) the most studied tool. Some applications of ANNs for pressure drop prediction include flow pattern recognition and pressure drop of two-phase flow \cite{Al-naser2016}, air-water pressure drop in vertical channels \cite{Najafi2021}, pressure drop in horizontal channels \cite{Barroso-maldonado2019},  pressure drop in fluidized beds \cite{Ghode2016}, pressure drop in venturi scrubbers \cite{Esmaeili2017}, pressure drop during condensation in inclined smooth tubes \cite{Zendehboudi2017}, and convective heat transfer and pressure drop in boiling and condensation \cite{Barroso-maldonado2019, Ledesma2018, Jose2017}. A common feature of these works is the excellent correlation degree reported by the ANN models. This feature motivates engineers and designers to implement these models in thermal design applications \cite{Yaji2022, Wang2020, Bhattacharyya2021}. However, these models are not easily reproducible for users, and their implementation is limited by the range of data used in training the ANN.

Transfer learning, which inspires the proposed model, is a novel tool used lately in situations where there is a lack of information related to a phenomenon. This technique uses a pre-trained ANN for an application related to the problem needed to be solved and using it as a base, retrain the pre-train ANN with a small dataset of the specific problem. It offers an increment in the performance of the ANN for the new problem. This methodology has been extensively explored in diverse fields, for instance, in data management \cite{Zhuang2021}, deep learning \cite{Tan2018}, chemistry \cite{Yamada2019}, diagnosis of machines \cite{Guo2019}, mechanical fault diagnosis \cite{Pan2019}. Similar to this concept, recently also have been proposed the physical-informed approach using deep learning model; where an ANN model is trained with small datasets, and it should respect a given set of physical laws. For instance, Raissi et al. \cite{Raissi2017} proved the paradigm of implicitly informing physical laws to a deep learning model in the solution of incompressible fluid flow, described by Navier-Stokes equations, encountered in the vortex shedding pattern in the cylinder wake (the Von Kármán vortex street). Additionally, the physical-informed neural networks approach has been used for the thermochemical simulation of a composite material \cite{AminiNiaki2021}. These novel techniques based on Artificial Intelligence (AI) obey the searching of models that represent complex phenomena governed by highly non-linear equations. In the case of the pressure drop of forced boiling flow, this phenomenon is currently modeled using empirical correlations; however, their accuracy problems still offer the opportunity for improvements. Whereas AI-based models address the problem of accuracy, they report problems of generality or extrapolability features.

In this context, this work presents a new paradigm for implementing an ANN for the two-phase flow pressure drop prediction on micro-channels. Transfer Learning inspires the paradigm since the proposed model is fed with the knowledge of an available correlation; this is like the pre-trained ANN mentioned above. The methodology can assist both: i) thermal analysis, rating, and sizing of heat exchangers used in low-temperature applications, ii) extend this kind of paradigm in other applications of transport phenomena, fluid mechanics, and refrigeration. The proposal paradigm is based on ANN, whose inputs are local quality, inner diameter, and pressure drop gradient fed from correlation. The output of the model is the frictional pressure drop. The document is organized as follows: Section 2 introduces the transfer learning concept and its application to the paradigm proposed in this paper; the transfer learning analogy applied in the modeling of transport phenomena is discussed in detail. Section 3 presents a literature survey of available correlations to predict pressure drop and the experimental database used for the ANN development; this section exposes the trend of the best correlations for pressure drop prediction and the need for more accurate models. Sections 4 and 5 report the CoINN development and the evaluation of its predictions, showing the methodology followed for the CoINN architecture election and the input correlation as the Transfer Learning inspired technique. Finally, the CoINN's success is demonstrated by the proving applicability to other scenarios not considered during training. The source code of the CoINN to predict the pressure-drop in micro-channels can be found on GitHub: \url{https://github.com/alejomonbar/CoINN}

\section{Transfer learning concept}

Transfer Learning (TL) is a concept of machine learning where a learned task can be reused in Machine Learning (ML) models development. It can be traced back to Lorine Pratt et al. \cite{Pratt1991}. TL has been beneficial for ML and recently developed widely in single Artificial Neural Networks (ANN) models and deep learning techniques. For these cases, TL offers the opportunity of working with a small dataset by reusing a pre-trained model. For example, in image classification, a model pre-trained with a large dataset of images and categories can specialize in a new dataset of a different category. 

The TL methodology can be viewed from the following point: the information usage provided from a single correlation to feed an ANN model; in this way, the ANN is enriched on the morphology of the physical phenomena. Hence, CoINN combines the correlation with other parameters to predict experimental results, instead of only comparing the correlation with an ANN model. This approach is proposed in terms of accuracy improvement, and ANN extrapolation features.

The nature of ANN is based on non-linear relationships because they are inspired by the behavior of biological neurons, specifically on the nervous system connections. The analogy to construct ANN is used to efficiently solve challenging problems found in engineering, managing provided data classified as inputs and targets. The data management consists of several steps including filtering data, pre-training, training, pre-validation, validation, and post-validation. During the training, similar to the synapses adjustment encountered in biological systems, the artificial synapses (identified in artificial computational sciences as weights) are represented by a set of parameters. They are mathematically adjusted using an algorithm to optimize the error in predictions. Their ability to model complex processes makes them promising tools for application in various areas of mechanical engineering, including transport phenomena, thermal sciences, and fluid dynamics \cite{Yaji2022, Zhuang2021}.

\begin{figure}[h]
\centering
\includegraphics[width=12cm]{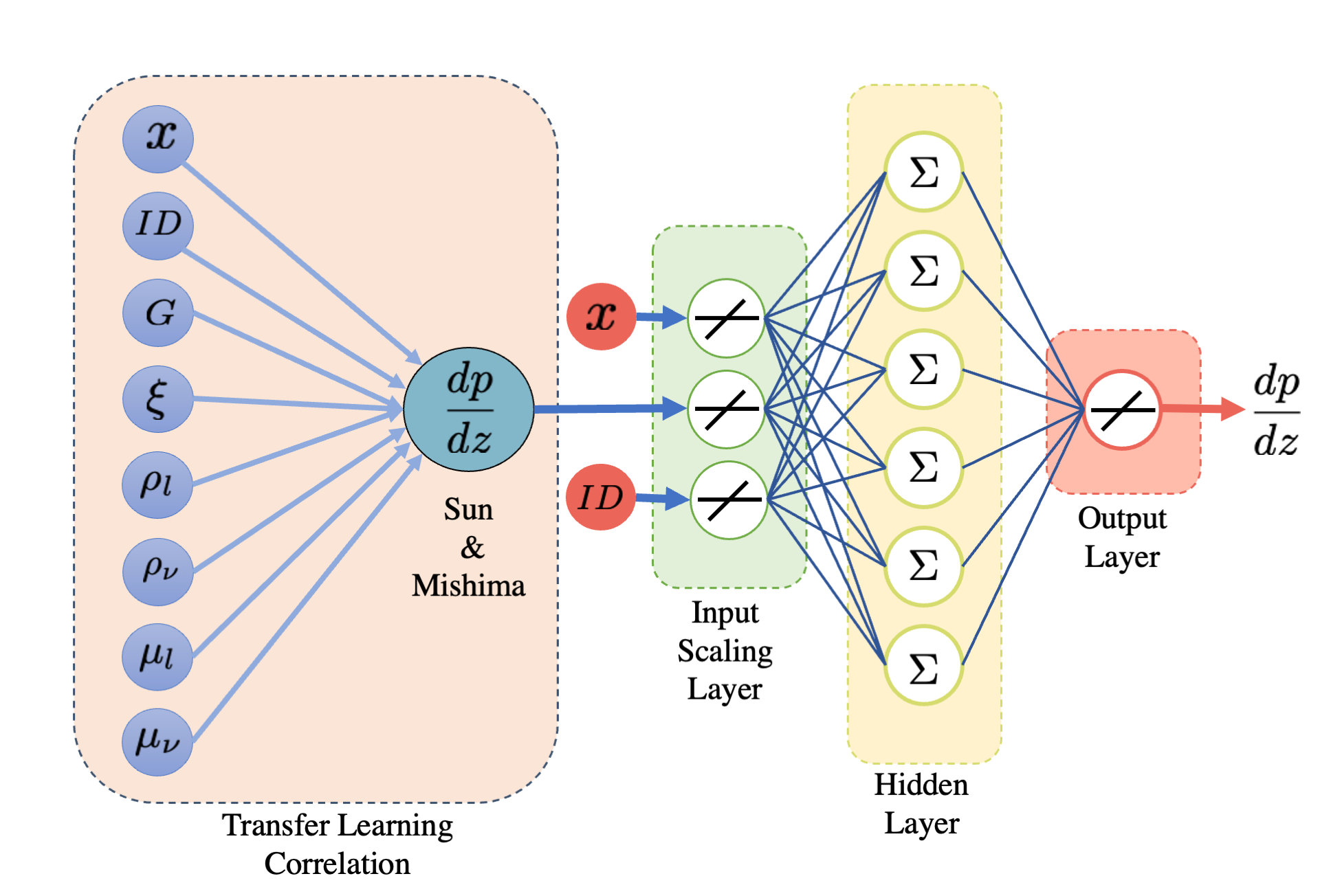}\\
\caption{\label{Fig1} CoINN architecture components (left) it is the transfer learning component, the Sun \& Mishima correlation, combined with the quality $x$ and the inner diameter {\it ID} are the inputs of the ANN; (right) it comprises an input scaling layer, a hidden layer with 6 neurons, and $tanh$ as activation function and an output scaling layer.}
\end{figure}

The ANN model used in this work consists of a multilayer feedforward network with three layers. The information flow begins when the input layer receives the input data and normalizes it. Next, a hidden layer with an activation function is responsible for gradually switching on/off the belonging back neurons. Finally, the information goes to the output layer to rescale the output information's data. Hence, each neuron acts as a processing element with simple input and output connections. In the case of hidden neurons, the incoming and outcoming information is being pondered with weights, $w_i$. The system first calculates the sum of the weight-weighted input elements and then applies an activation function to obtain the output value (y); a commonly used activation function is hyperbolic tangent. Fig. \ref{Fig1} shows the architecture proposed in this study for estimating friction pressure drop, and the flow information throughout the whole neural network. The first part involves the determination of the Sun \& Mishima pressure drop, which is considered the transfer learning step, then, the pressure drop calculated with Sun \& Mishima, the inner diameter, and the quality are passed to an ANN as inputs. Finally, the training of the ANN adjusts the weights to produce results close to those experimentally obtained. To measure the accuracy of the proposed model, we use the mean relative error ($mre$) expressed in Eq. (\ref{Eq:mre}),

\begin{equation}
mre[\%] = 100\left(\frac{1}{N}\sum_{i=1}^N \frac{|t_i - y_i|}{|t_i|}\right)
\label{Eq:mre}
\end{equation}

where $t_i$ is an experimental point, $y_i$ is the output of the ANN for configuration $t_i$, and $N$ is the number of experimental points.

\section{Proposed method}
\subsection{Correlations survey}
The decision to use Sun and Mishima comes from an extensive literature review of the correlations that are currently available to calculate the pressure drop in a two-phase flow. Below is exposed the study of three different correlations that showed the lowest Absolute Average Deviation (AAD) compared with experimental data from Barraza et al. \cite{Barraza2016}. They are the Awad and Muzynchka \cite{Awad2008}, Sun and Mishima \cite{Sun2009}, and Ciccitti et al. \cite{Cicchitti1959}, with AAD of 17\%, 13\%, and 18\%, respectively. Additionally, these correlations are evaluated with the mean relative error ($mre$) metric for the same dataset in Barroso et al. \cite{Barroso-maldonado2019}, getting $mre$ of 23.9\% (Sun and Mishima), 24.1\% (Awad and Muzychka), and 25.3\% (Cicchitti et al.). Ciccitti et al. and Awad and Muzychka assume homogeneous flows from these correlations, while Sun and Mishima assume separated phase flows. The Homogeneous model is expressed according to Eq. (\ref{Eq2}).

\begin{equation}
\left(\frac{dp}{dz}\right)_{2ph} = 2 f_{2ph}\frac{G^2}{ID\rho_{mix}}
\label{Eq2}
\end{equation}

Here, the two-phase flow is treated as a single fluid combining the characteristics of the two phases. The friction factor is calculated with the Churchill correlation \cite{Awad2008} for micro-channels, as shown in Eq. (\ref{Eq3}). 

\begin{equation}
f_{2ph} = 2 \left[ \left(\frac{8}{Re}\right)^{12} + \frac{1}{(a + b)^{1.5}}\right]^{\frac{1}{12}}
\label{Eq3}
\end{equation}

Where $a$ and $b$ are defined in Eqs. (\ref{Eq4}) and (\ref{Eq5}).

\begin{equation}
a = \left[ 2.457\left(\frac{1}{\left(\frac{7}{Re}\right)^{0.9} + 0.27\xi}\right)\right]^{16}
\label{Eq4}
\end{equation}

\begin{equation}
b = \left(\frac{37530}{Re}\right)^{16}
\label{Eq5}
\end{equation}

The Reynolds number presented in the parameter $b$ is calculated using Eq. (\ref{Eq6}).

\begin{equation}
Re = \frac{G {\it ID}}{\mu_{2ph}}
\label{Eq6}
\end{equation}

The Awad and Muzychka correlation differs from the Chicitti et al. in calculating the viscosity, for the Awad and Muzychka correlation presented in Eq. (\ref{Eq7}) is used.

\begin{equation}
\mu_{2ph} = \mu_l \frac{2\mu_l + \mu_v - 2 (\mu_l - \mu_v) x}{2\mu_l + \mu_v + (\mu_l - \mu_v) x}
\label{Eq7}
\end{equation}

While for the case of Chichitti et al., Eq. (\ref{Eq8}) is proposed.

\begin{equation}
\mu_{2ph} = (1 - x)\mu_l + x \mu_v
\label{Eq8}
\end{equation}

In the case of the Sun and Mishima correlation, the model distinguishes the liquid and the vapor phase as separate flows. Here, it is taken the flow in the liquid section Eq. (\ref{Eq9}) with a correction factor $\phi_l$ between the two-phase flow Eq. (\ref{Eq10}). 

\begin{equation}
\left(\frac{dp}{dz}\right)_{2ph} = \left(\frac{dp}{dz}\right)_{l} \phi_l^2 = f_l \frac{G(1 - x)^2}{2 {\it ID}\rho_l}\phi_l^2
\label{Eq9}
\end{equation}

\begin{equation}
\phi_l^2 = 1 - \frac{C}{X^{1.19}} + \frac{1}{X^2}
\label{Eq10}
\end{equation}

Where X is the relation between the pressure drop for the liquid and vapor using Eq. (\ref{Eq11}).

\begin{equation}
X = \frac{\left(\frac{dp}{dz}\right)_l}{\left(\frac{dp}{dz}\right)_v}
\label{Eq11}
\end{equation}

The constant C changes depending on the flow conditions; the constant C is calculated with Eq. (\ref{Eq12}) for laminar flow. 

\begin{equation}
C = 26 \left(1 + \frac{Re_l}{1000}\right) \left(1 - \textrm{exp}\left(-\frac{0.153}{0.27L_a + 0.8}\right)\right)
\label{Eq12}
\end{equation}

Where La is the Laplace number, this factor considers the surface tension effects and it can be computed with Eq. (\ref{Eq13}).

\begin{equation}
L_a = \frac{1}{D_H}\sqrt{\frac{\sigma}{g(\rho_l - \rho_v)}}
\label{Eq13}
\end{equation}

For the case of turbulent flow, the constant C is determined using Eq. (\ref{Eq14}).

\begin{equation}
C = 1.79\left(\frac{Re_v}{Re_l}\right)^{0.4}\left(\frac{1-x}{x}\right)^{0.5}
\label{Eq14}
\end{equation}

\subsection{Database}
\begin{figure}[h]
\centering
\includegraphics[width=14cm]{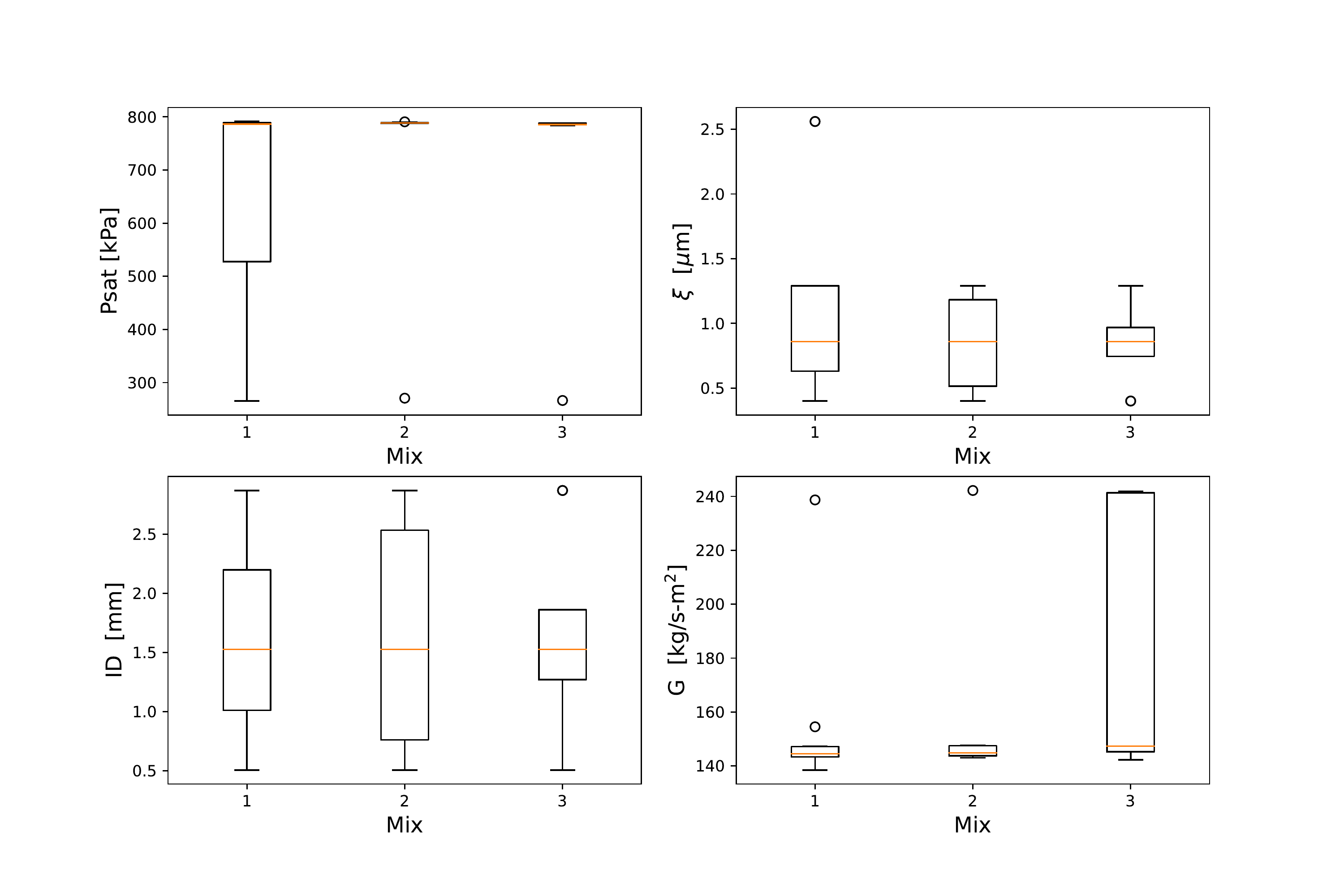}\\
\caption{\label{FigBox} Box plot that represents the distribution of the different characteristics mentioned in Table \ref{Tab1} for the three mixtures of hydrocarbon, synthetic, and binary mixtures with labels 1, 2, and 3, respectively. The dots represent outlier points.}
\end{figure}

\begin{figure}[h]
\centering
\includegraphics[width=12cm]{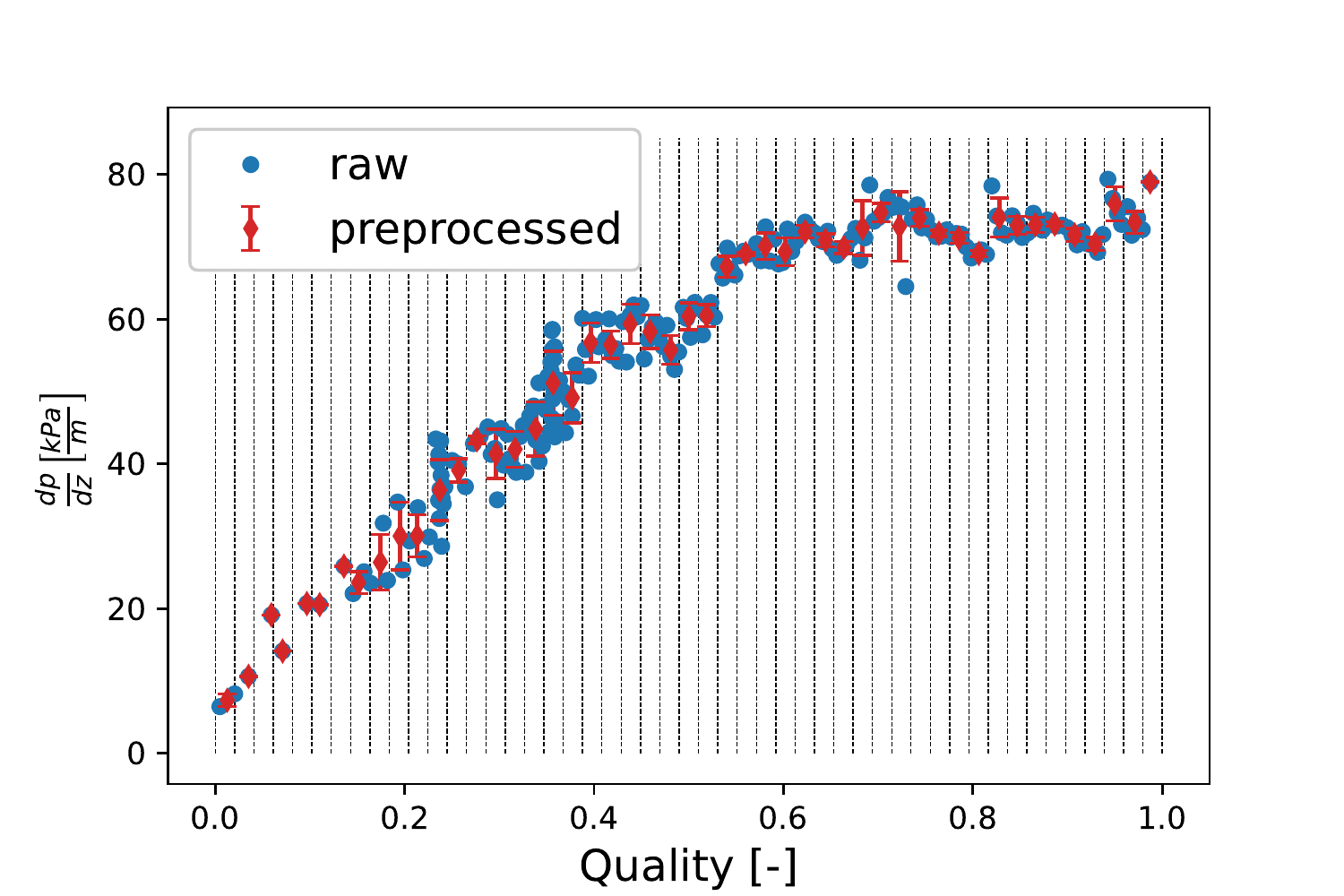}\\
\caption{\label{FigRaw} Preprocessing of the Barraza et al. data for experiment 1. To reduce the noise the quality is divided into 50 equally spaced regions (dashed vertical lines) and the experimental points (blue-circles) that fall in each region are averaged to get the preprocessed points (red-diamonds). The error bars indicates the standard deviation of the points that falls in the region.}
\end{figure}

In this section, we describe the experimental data used in this work. The main dataset, from Barraza et al. \cite{Barraza2016}, describes the frictional pressure behavior in boiling zeotropic mixtures in horizontal tubes with circular sections. There are three sets of working fluids; the first is composed of hydrocarbon mixtures consisting of methane, ethane, and propane, and in some cases with the addition of nitrogen. The second is composed of synthetic refrigerants R-14, R-23, R-32, R-134a, and Ar with different molar compositions, and the last set is composed of a binary mixture of methane and ethane. The inner diameter of the various experiments ranges from 0.5 mm to 2.9 mm, roughness from 0.4 $\mu$m to 2.56 $\mu$m, critical pressure from 265 kPa to 789 kPa, and mass flux from 143 to 242 kg/s-m$^2$. The set of 37 experiments, summarized in Table \ref{Tab1}, includes 15 hydrocarbon mixtures, 14 synthetic mixtures, and 8 binary mixtures. Fig. \ref{FigBox} shows the distribution of the pressure, mass flux, ID, and $\xi$ for the three mixture sets. For more details about the experiments see Ref. \cite{Barraza2016}. The raw data of the 37 experiments includes 9524 individual points. This information is preprocessed by dividing the vapor quality into 50 equally spaced regions. The points that fall inside each region are averaged to eliminate part of the measurement noise. Fig. \ref{FigRaw} shows this procedure in experiment 1, the raw data points are the blue circles and the preprocessed data the red diamonds with the standard deviation of the experiments shown in the error bars. After the preprocessing, the data is reduced to 1565 experimental points. The experiments are used to train the ANN and to validate that there is no overfitting in the ANN.

\setlength{\tabcolsep}{5pt}
\begin{table*}[h]
\center
\caption{Barraza et al. range of conditions for the 37 experiments was divided into three categories: hydrocarbons mixtures, synthetic mixtures, and binary mixtures.}
\begin{tabular}{|p{0.23\textwidth}|p{0.16\textwidth}|p{0.13\textwidth}|p{0.13\textwidth}|p{0.09\textwidth}|p{0.09\textwidth}|}
\hline
Mixtures &Composition & Pressure [kPa] &  Mass Flux  [kg/s-m$^2$] & {\it ID}  [mm] & $\xi$ [$\mu$m]\\ 
\hline 
CH$_4$/C$_2$H$_6$/C$_3$H$_8$/N$_2$ & 0.27-0.45/ 0.21-0.35/ 0.12-0.2/ 0.0-0.4 & 265-789 & 143-242 & 0.5-1.5 & 0.4-2.56 \\ 
\hline 
R-14/R-23/R-32/\par R-134a/Ar &0.21-0.35/ 0.09-0.15/ 0.09-0.15/ 0.21-0.35/ 0.0-0.4 & 271-790&143-242&0.5-1.5&0.4-1.29\\
\hline
CH$_4$/C$_2$H$_6$&0.4/0.6&266-789&145-242&0.5-2.9	&0.4-1.29\\ \hline
\end{tabular}
\label{Tab1}
\end{table*}

\subsection{Model development}

\begin{figure}[h]
\centering
\includegraphics[width=10cm]{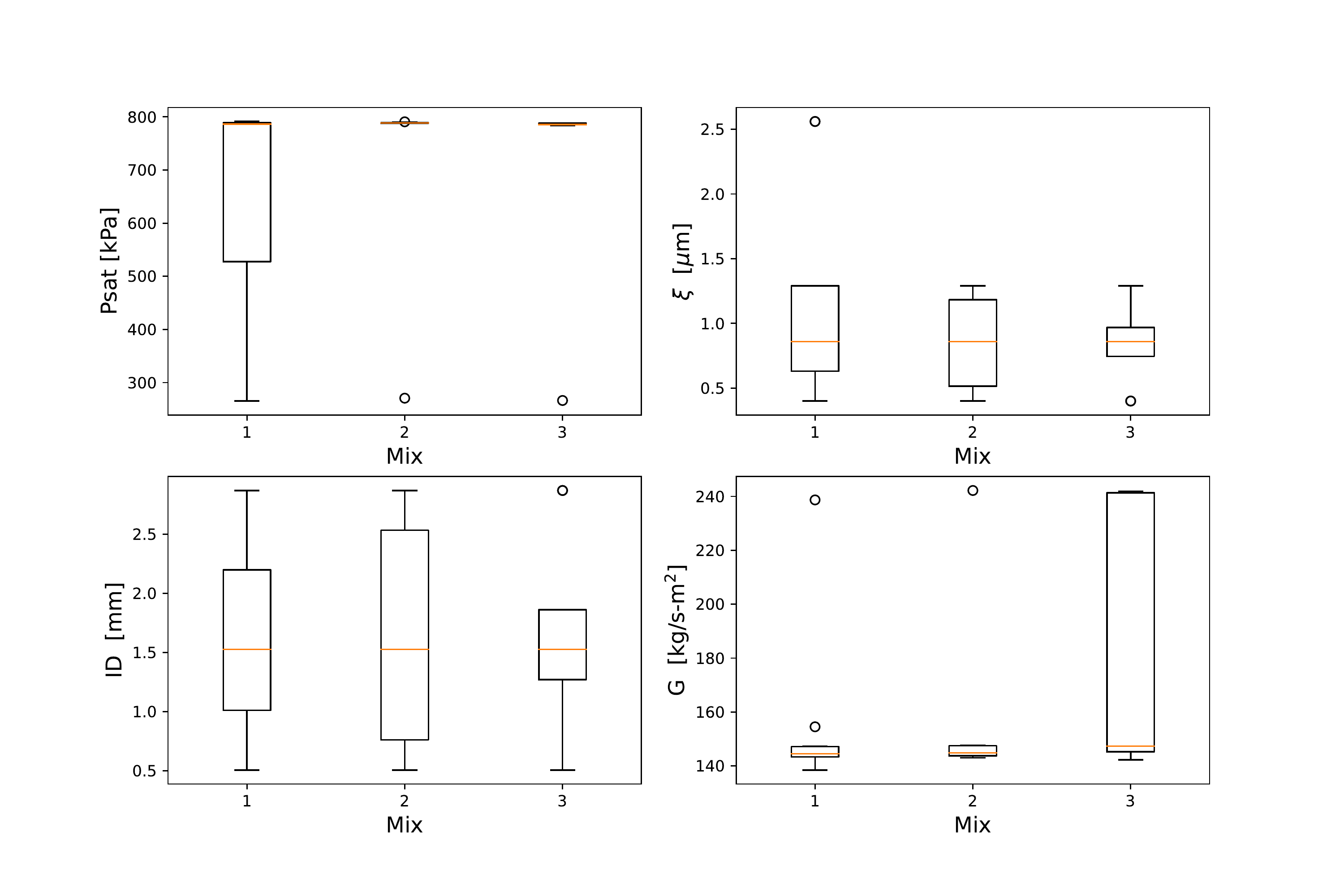}\\
\caption{\label{Fig2} Correlation diagram of the different direct experimental measurements and the experimental pressure drop (Exp). For this case, a PCC is used for the diagram}
\end{figure}

One research opportunity found in the literature is the absence of a concrete methodology to select independent variables that govern the pressure drop of a single-, two-, or multiple phases. Researchers are then faced with this challenge, and this work is no exception. In Fig. \ref{Fig2}, a correlation coefficient diagram is presented to show the dependency of pressure drop on several common experimental variables that have been proposed. The set of variables can be identified by those reporting geometry information (roughness and inner diameter), mixing composition, and mass flux; but also by those featuring the flow such as the homogenous Reynolds number, the liquid Reynolds number, the vapor Reynolds number, the liquid frictional factor, and the vapor frictional factor. Such a set of variables shows a correlation level with pressure drop. The correlation coefficient diagram in Fig. \ref{Fig2} shows the normalized covariance of two different variables. This method looks for the influence degree of a set of variables in experimental pressure drop. The result of this figure is scaled to a blue and red color representing -1 and 1, respectively. This analysis was assisted by the Pearson correlation coefficient (PCC) \cite{Hotelling1953}, see Eq. (\ref{Eq15}). This statistic measures the linear relation between two datasets.

\begin{equation}
\rho_{x, y} = \frac{cov(x, y)}{\sigma_x\sigma_y}
\label{Eq15}
\end{equation}

Where {\it cov} is the covariance, $\sigma_x$ and $\sigma_y$ are the standard deviation of dataset $x$ and $y$, respectively. PCC was used when developing this figure since the investigating variables have easy insight and are directly measurable. 

From Fig. \ref{Fig2}, it is graphically noticed a diagonal matrix on which a lecture of an x-variable to a y-variable is gotten. For instance, taking the experimental pressure drop (Exp) on the horizontal axis, its vertical positions represent the degree of correlating it with other inputs. From the results, roughness ($\xi$) with 0.6 and inner diameter ({\it ID}) with -0.8 are the most correlated variables to the pressure drop. However, notice that the {\it ID} and the $\xi$  have a correlation coefficient of +0.7, meaning they are highly correlated and change similarly in the experiments. 

A similar study is performed to those featuring the fluid flow, such as mixture Reynolds number, liquid Reynolds number, vapor Reynolds number, liquid friction factor, and vapor friction factor. These parameters are obtained using measurable information, and hence a linear relationship is not ensured. For this reason, to analyze their degree of correlation with experimental pressure drop is proposed the Spearman rank-order correlation coefficient \cite{Stephanou2021}; such a method measures the monotonicity between two datasets. These coefficients give a sense of how variables that change through the quality range are related.

\begin{figure}[h]
\centering
\includegraphics[width=10cm]{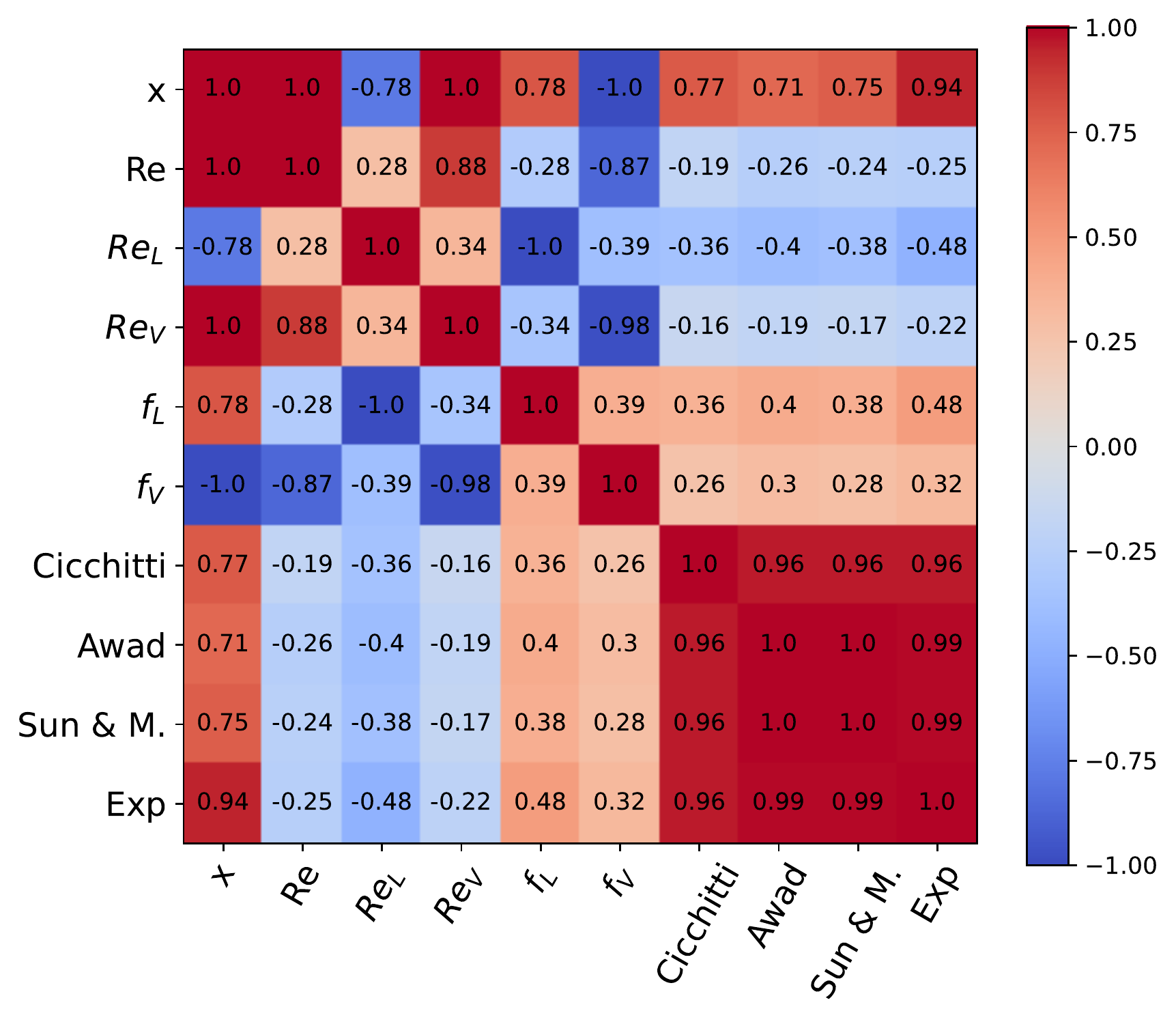}\\
\caption{\label{Fig3} Correlation diagram of different parameters computed using experimental measurements. For this case, a Spearman rank-order correlation coefficient is used for the diagram.}
\end{figure}

From Fig. \ref{Fig3}, it is selected the Sun \& M. correlation and the vapor quality $x$ with a correlation coefficient of +0.99 and +0.94, respectively. Hence, we choose ID, Sun \& Mishima correlation, and the vapor quality as input variables and compare them with a model that includes all the variables shown in both diagrams in Fig. \ref{Fig2} and Fig. \ref{Fig3} to see if they are enough to produce acceptable results. 

\section{Results and Comparison}

Once the input features of CoINN are carefully decided, the next is to evaluate the model's sensitivity to hidden layer complexity; hence, exploring the effect of hidden neurons number on the ANN mapping abilities is necessary. The followed procedure consists of four steps: training step, validation step, testing step, and post-validation step. The selected data set comes from Barraza et al. \cite{Barraza2016}, outing 32 of 37 mixtures for the training step. The remaining 5 mixtures, which correspond to 238 experimental samples used as a post-validation test. These 32 mixtures are composed of 1327 experimental points divided randomly to get: 70\% for the training step, 15\% for the validation step, and 15\% for the testing step. Here, the number of hidden neurons increases from 1 to 15, generating in each step a different network architecture; in each network, the training process initializes testing 1000 random initial solutions. For each of the 1000 initial solutions, using the mean square error (mse) as the cost function, we use the Levenberg-Marquardt algorithm during 1000 iterations to find the weights that minimize the cost function. Once, we have the 1000 solutions with their optimal weights, we select the solution with the best performance in terms of minimum mre.

\begin{figure}[h]
\centering
\includegraphics[width=12cm]{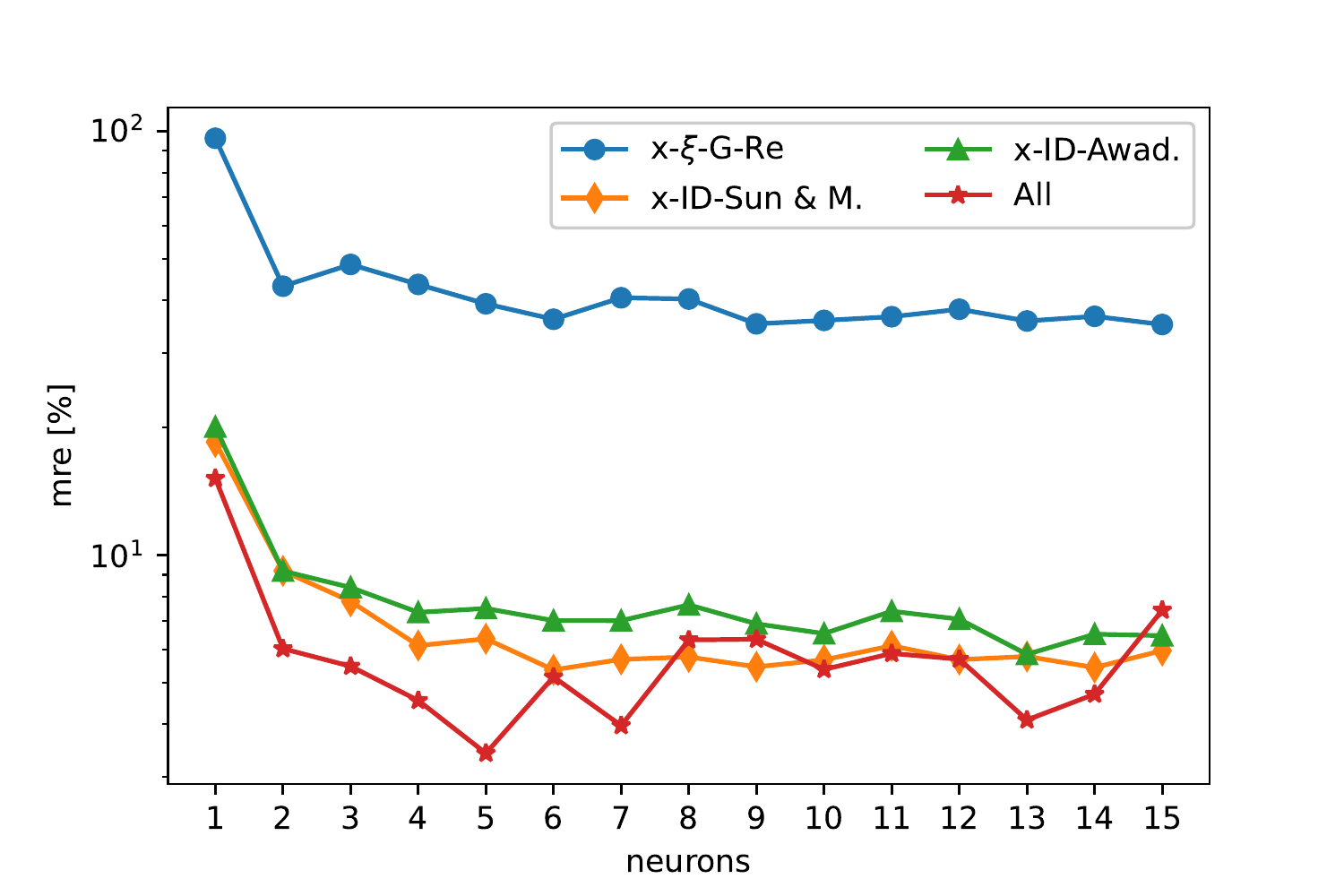}\\
\caption{\label{Fig4} Four different ANN models are proposed to predict the pressure drop; we quantify the performance of each method for different neurons in the hidden layer using the $mre$ of the 37 experiments from Barraza et al. \cite{Barraza2016}}
\end{figure}

Fig. \ref{Fig4} shows the ANN sensibility in its predictions vs its complexity having low or high neurons number in the hidden layer. This study assists in defining the most appropriated architecture achieving two main features: simplicity for users and accuracy, both elements are simultaneously considered to avoid overfitting problems. The vertical axis reports the averaged $mre$ from the test and the post-validation sets. The horizontal axis presents the neurons' number on the hidden layer. This study was performed for the three proposed input configurations: i) quality, roughness, mass flux, and Reynolds (blue circle line); ii) quality, inner diameter, Sun \& Mishima (orange diamond line); iii) quality, inner diameter, Awad (green triangle line); and iv) all the possible variables (shown in Fig. \ref{Fig4}, red-star line). 

Several important observations can be highlighted from the results shown in Fig. \ref{Fig4}, e.g., when using directly measurable parameters and the Reynolds number, the results offer an opportunity for considerable improvement. This inspires the analogous usage of the transfer learning technique. The error curves fall drastically when including the current correlations such as Sun \& Mishima and Awad in conjunction with other representative flow parameters (see green-triangle and orange-diamond lines). The red-star line represents the case when all variables are considered. This line does not differ considerably from Sun \& Mishima and Awad, despite being more robust when considering all the variables included in the horizontal axis of Fig. \ref{Fig2} and Fig. \ref{Fig3}. Using this discussion, it is proposed to use the model whose inputs are quality, inner diameter, and Sun \& Mishima (x-ID-S\&M). On the other hand, the horizontal axis in Fig. \ref{Fig4} represents the hidden neurons and the sensitivity model to this parameter. From Fig. \ref{Fig4}, it can be concluded that the computed $mre$ (5.3\%) does not decrease significantly after six neurons. Therefore, we keep 6 neurons in our ANN hidden layer model and avoid overfitting problems with this selection.

\begin{figure}[h]
\centering
\includegraphics[width=12cm]{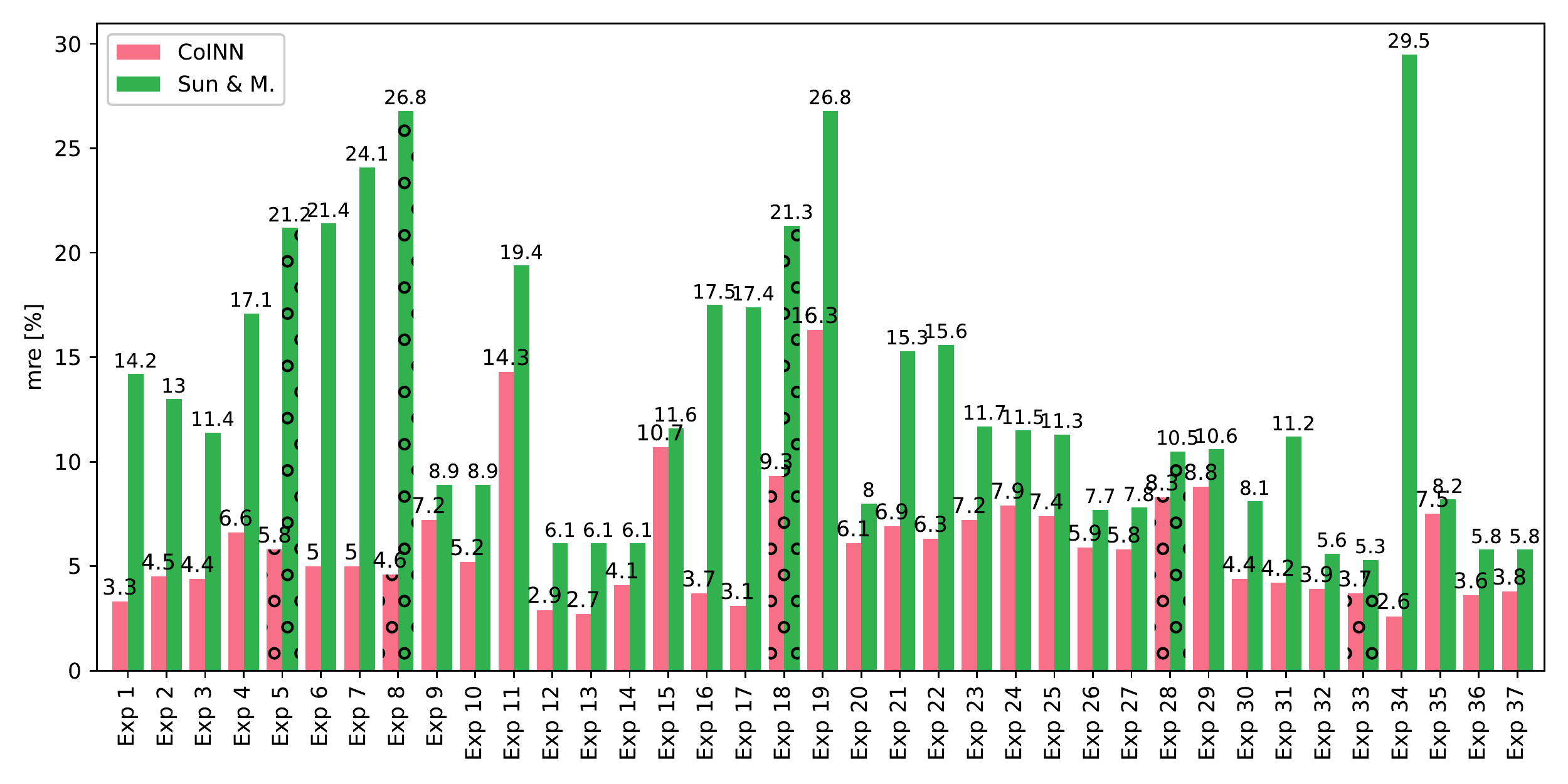}\\
\caption{\label{Fig5} Results of CoINN vs the Sun \& Mishima correlation $mre$. The experiments with circles are those not used during ANN training.}
\end{figure}

A detailed analysis of the ANN chosen (x-ID-Sun \& M model with 6 hidden neurons) is depicted in Fig. \ref{Fig5}. The y-axis represents the reported $mre$ by each experiment configuration (each configuration is composed of several experiments run from the dew to bubble saturation point). In contrast, the x-axis represents such  an experimental configuration. The reported results show the ANN outperforming the single Sun \& Mishima correlation for every experimental configuration, including those not included in training (experiments 5, 8, 18, 28, and 33). According to final global results, the conventional model predicted pressure drop with averaged $mre$ of 13.3\%, while the ANN model predicted with averaged $mre$ of 6.1\%. 

\begin{figure}[h]
\centering
\includegraphics[width=7cm]{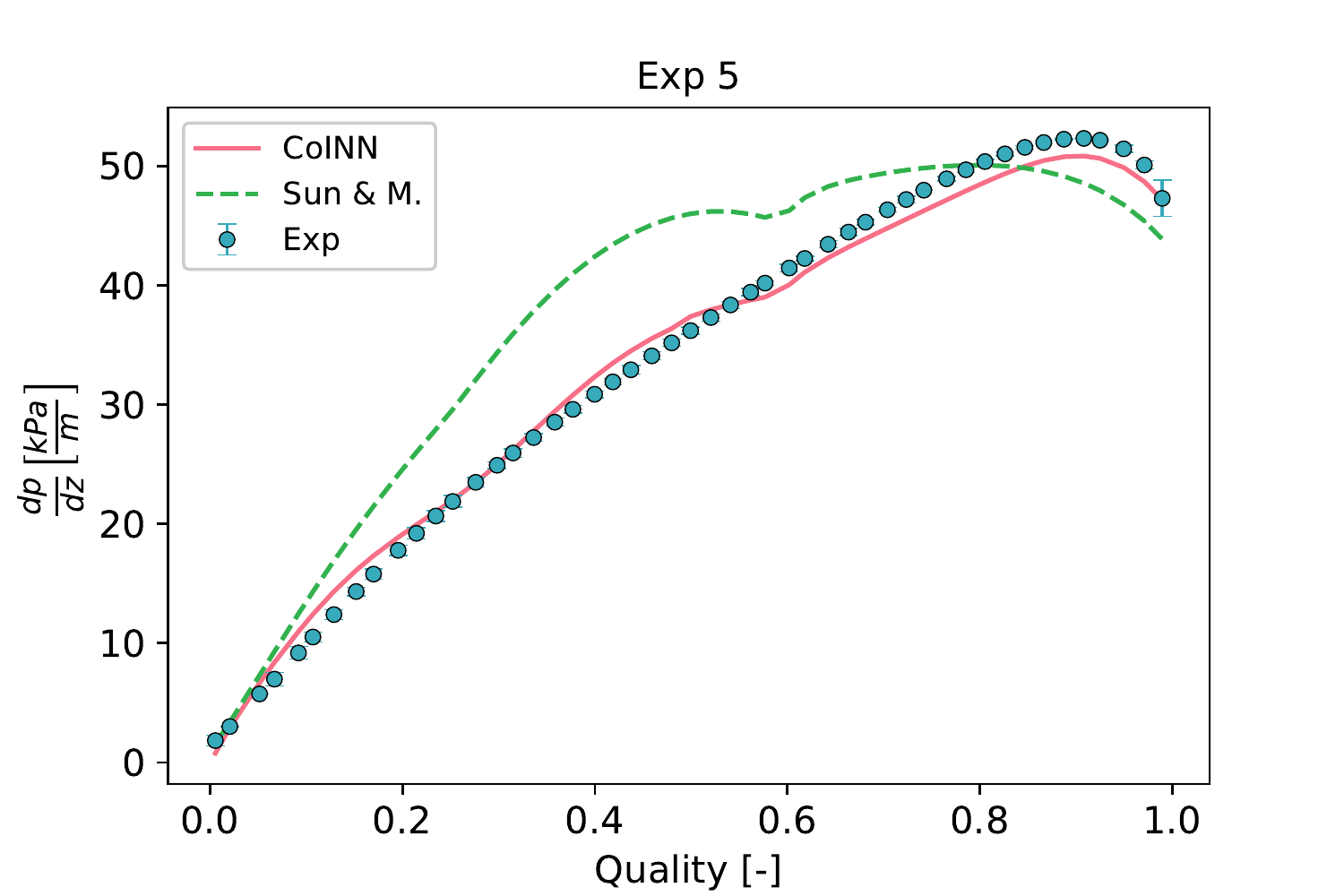}
\includegraphics[width=7cm]{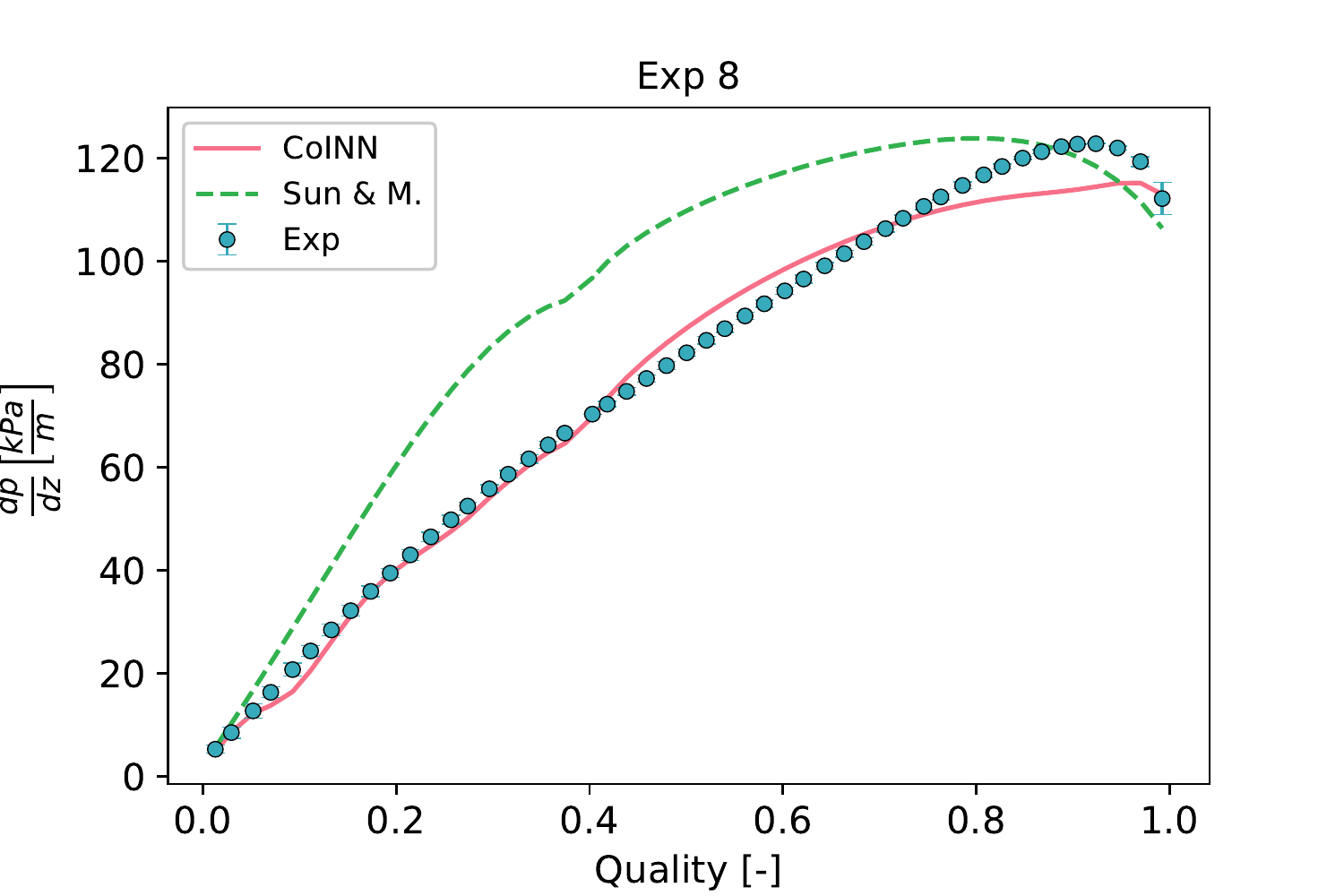}
\caption{\label{Fig6} Experimental results (circles), Sun \& Mishima correlation (dashed-green), and CoINN (red-solid) for the hydrocarbon mixtures not used for the training of the ANN. These experiments are composed by methane (45\%), ethane (35\%), and propane (20\%) both differ in roughness 2.56 (left) and 0.86 $\mu$m (right) and in mass flux 139 kg/s-m$^2$, and 239 kg/s-m$^2$.}
\end{figure}

\begin{figure}[h]
\centering
\includegraphics[width=7cm]{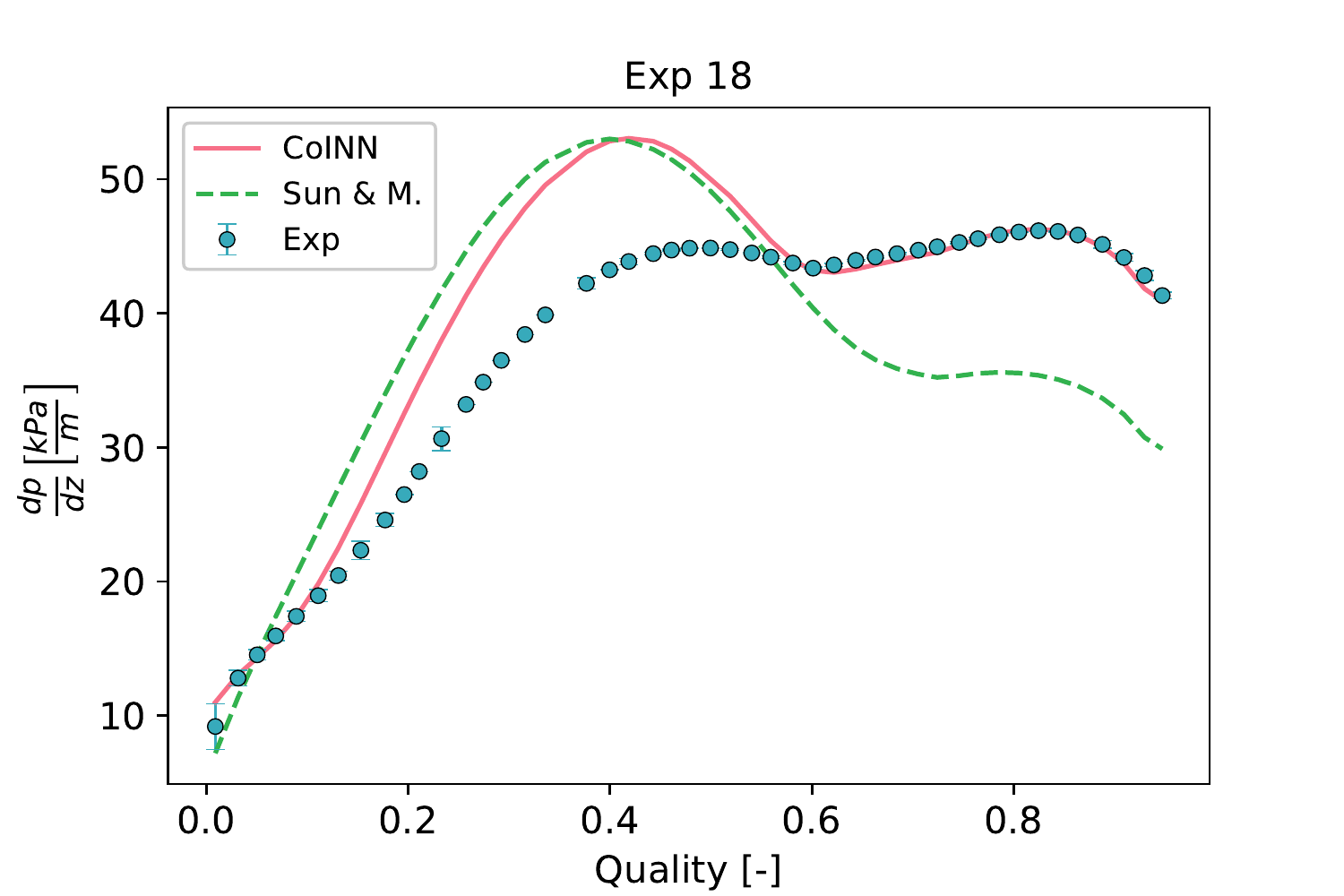}
\includegraphics[width=7cm]{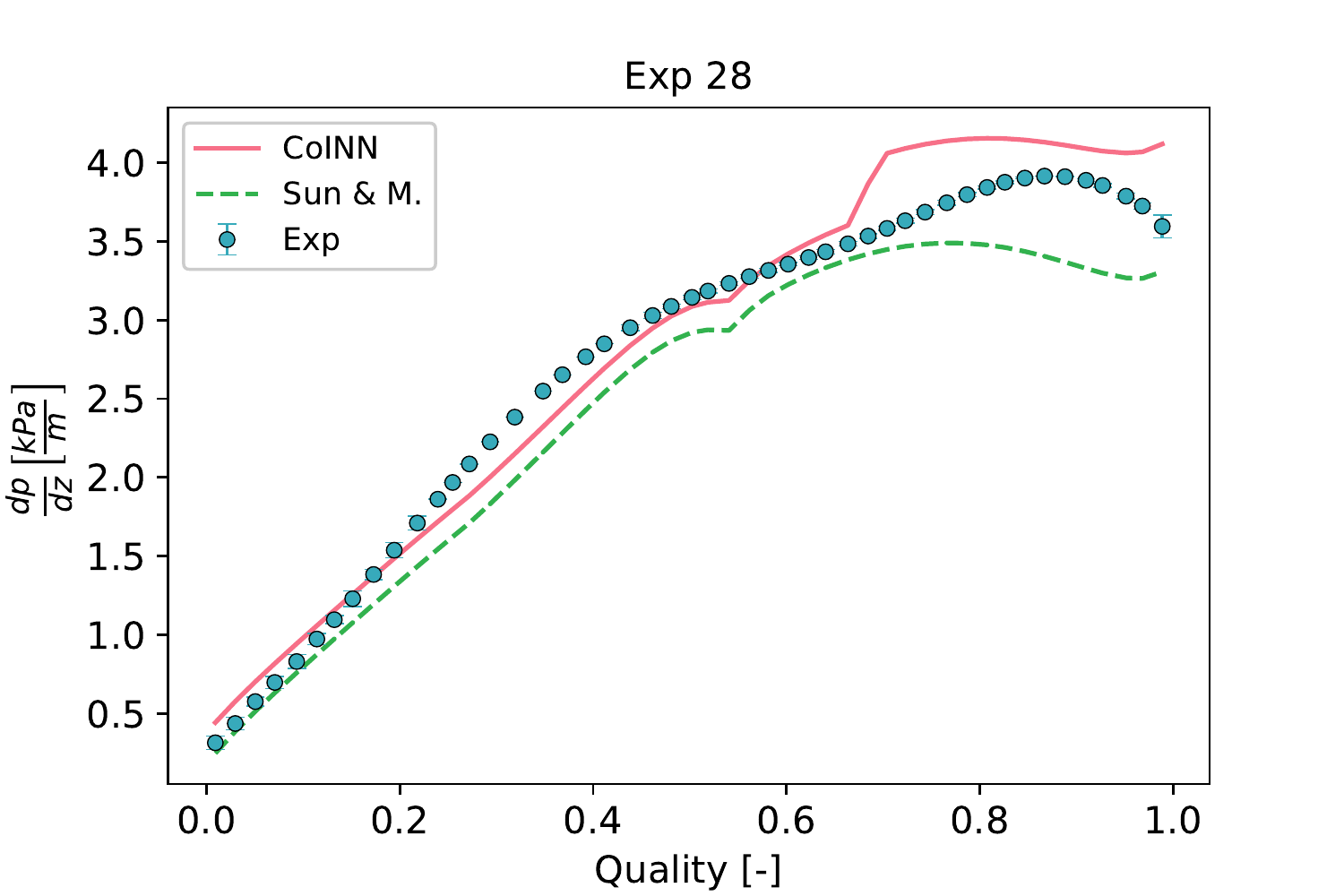}
\caption{\label{Fig7} Experimental results (circles), Sun \& Mishima correlation (dashed-green), and CoINN (red-solid) for the synthetic mixtures not used in the ANN training. These experiments are composed by R-14 (28\%), R-23 (12\%), R-32 (12\%), R-134a (28\%), both differ in ID 0.56 (left) and 2.87 mm (right) and in surface roughness 1.29 (left) and 0.4 $\mu$m (right).}
\end{figure}

\begin{figure}[h]
\centering
\includegraphics[width=12cm]{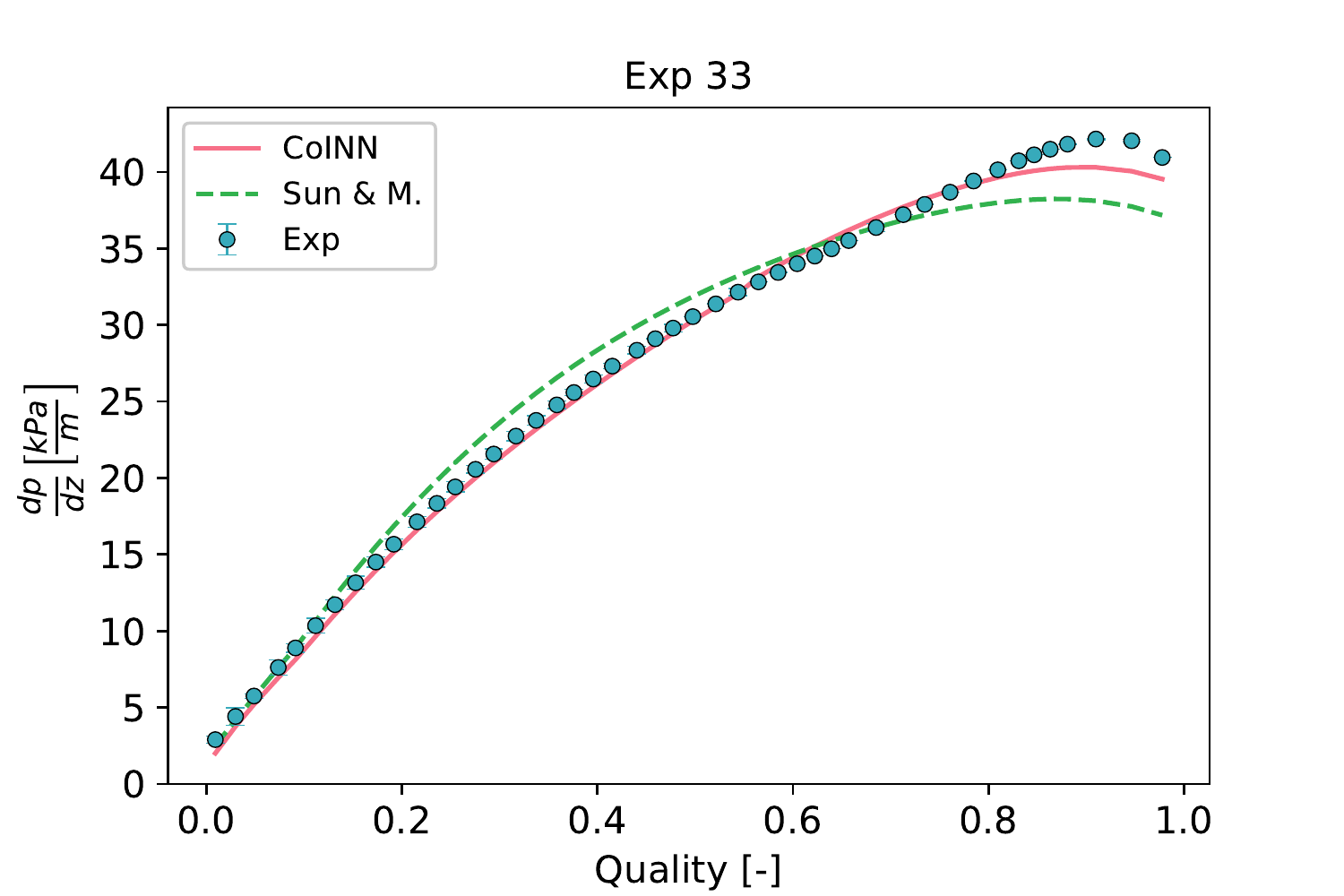}
\caption{\label{Fig8} Experimental results (circles), Sun \& Mishima correlation (dashed-green), and CoINN (red-solid) for the binary mixtures not used for the training of the ANN. These experiments are composed of methane (40\%) and ethane (60\%).}
\end{figure}

The experiments not used in the training of CoINN, with circles in Fig. \ref{Fig5}, are shown in detail in Figs. \ref{Fig6}-\ref{Fig8}. In these and subsequent figures, the pressure drop as a quantity is denoted by $dp/dz$ in the y-axis. Fig. \ref{Fig6} shows the results of experiments 5 (left) and 8 (right); these experiments have a composition of methane (45\%), ethane (35\%), and propane (20\%), an inner diameter of 1.5 mm, evaporating pressure of 268 kPa, with a variation in their roughness 2.56 and 0.86 $\mu$m and their mass flux 139 and 239 kg/s-m$^2$, respectively.  According to this figure, CoINN improves considerably through the whole range of quality. This is quantified with a $mre$ of 5.8\% and 4.6\% compared with 21.2\% and 26.8\% of the Sun \& Mishima correlation, respectively. Fig. \ref{Fig7} shows experiment 18 (left) and 28 (right), these experiments have synthetic mixtures of R-14 (28\%), R-23 (12\%), R-32 (12\%), R-134a (28\%), and argon (20\%), with evaporating pressure 788 kPa, mass flux 144 kg/s-m$^2$, and inner diameter 0.56 and 2.87 mm and surface roughness 1.29 and 0.4 $\mu$m, respectively. In experiment 18 there is a big improvement for the prediction in quality range 0-0.1 and 0.6-1.0 compared to the Sun \& Mishima correlation. Experiment 28 shows improvement through the quality range of 0 to 0.6, but from there the tendency is not adjusted to that of the experiment. However, the same it is observed for the Sun \& Mishima case. Lastly, Fig. 8 shows experiment 33. This experiment is a binary mixture of methane (40\%) and ethane (60\%), with an inner diameter of 1.527 mm, a roughness of 0.86 $\mu$m, evaporating pressure of 788 kPa, and mass flux 242 kg/s-m$^2$. For this experiment, both models have low $mre$, but the ANN fits better to the experiment; this shows the power of the model proposed, using as a reference the data gathered from the correlation and improved by the training on the experimental data.

To get an insight into the proposed model extensibility, we test CoINN against experimental data under different experimental configurations not used during the training of CoINN. In this case, we compare the CoINN pressure drop predictions vs the Keniar et al. \cite{Keniar2021} experimental results. The authors reported the behavior of frictional pressure drop in condensation of zeotropic mixtures in horizontal tubes with circular and square sections. The experiments were conducted using R-134a, R-245fa, and R-1234ze(E). The frictional pressure drop was evaluated under different saturation pressures, ranging from 178 kPa to 1318 kPa; the mass flux range between 50 to 200 kg/s-m$^2$, the tube segment of 240 mm long, the internal diameter of 1.55 mm, and roughness of 0.5 $\mu$m. These comparisons corroborate the capabilities of the proposed model to predict pressure drop on a new set of configurations not used for the CoINN training. The results are shown in figures 9-11 where three working fluids are tested R134a, R245fa, and R1234ze(E). For the case of R-245fa and R1234ze(E), they are new working fluids for the ANN. Additionally, the experimental conditions differ from those used for training. For the saturation pressure, the range for training is between 265 to 790 kPa, and the Keniar et al. between 178.1 to 1317.9 kPa which is outside the range of those data points used for training. 

\begin{figure}[h]
\centering
\includegraphics[width=12cm]{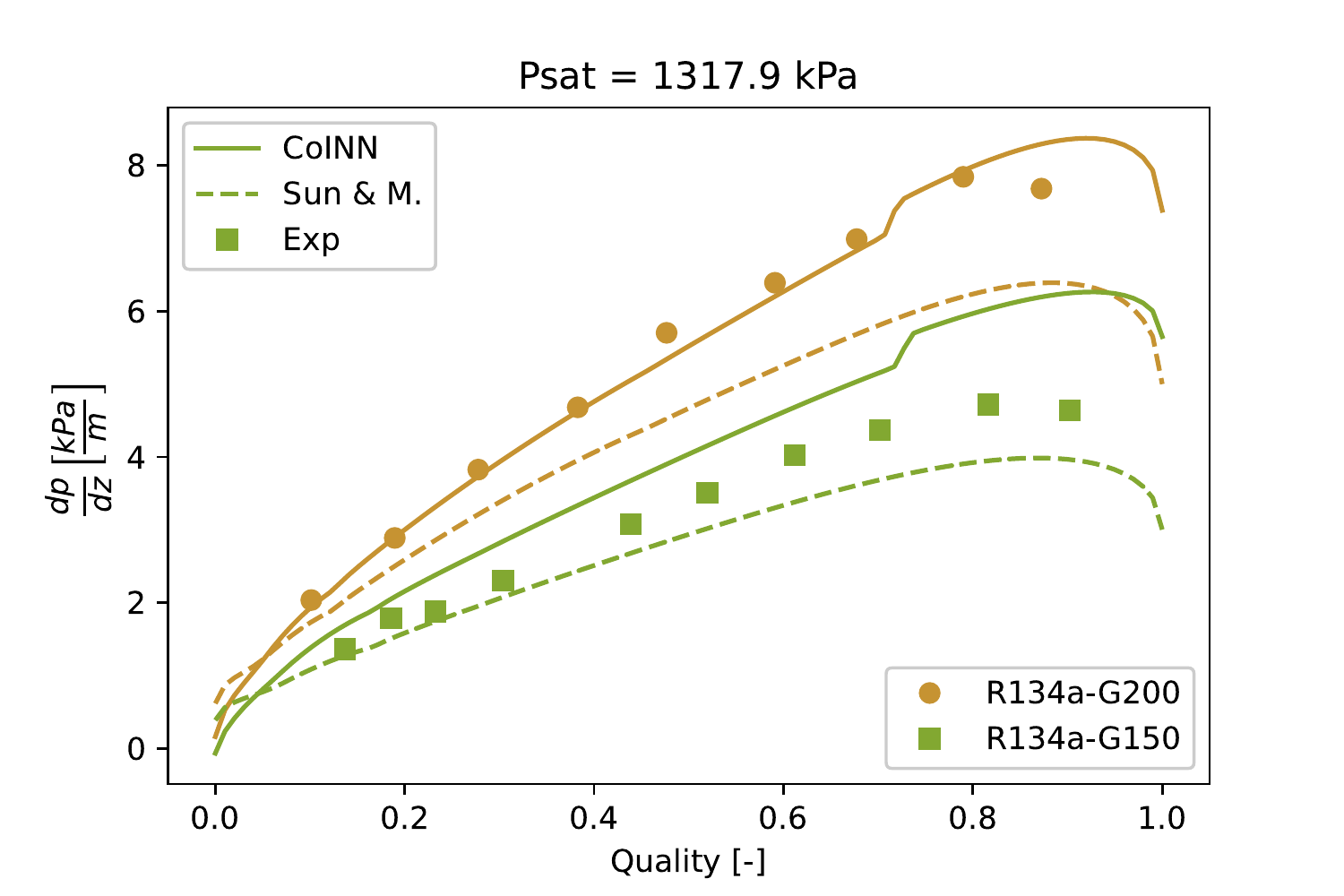}
\caption{\label{Fig9} Results comparing CoINN, the Sun \& Mishima correlation, and the experimental results for R134a. The solid lines represent the ANN solution, the dashed lines the Sun \& Mishima solution, and the markers circles for the experimental results with a mass flux of 200 kg/s-m$^2$ and squares for the experimental results with a mass flux of 150 kg/s-m$^2$.}
\end{figure}

Fig. \ref{Fig9} shows the results of CoINN (solid line), the Sun \& Mishima correlation (dashed line), and experiments with circles for a mass flux of 200 kg/s-m$^2$ and squares for 150 kg/s-m$^2$ with working fluid R134a and saturation pressure 1317.9 kPa. Even though this composition is used for training, the pressure drop of 1317.9 kPa is much greater than the maximum of the experiments, 790 kPa. The $mre$ for the mass flux of 150 kg/s-m$^2$ is 22.3 and 12.8\% for CoINN and the Sun \& Mishima correlation, respectively. For the mass flux of 200 kg/s-m$^2$, the $mre$ are 3.2\% and 17.3\%, respectively.

\begin{figure}[h]
\centering
\includegraphics[width=12cm]{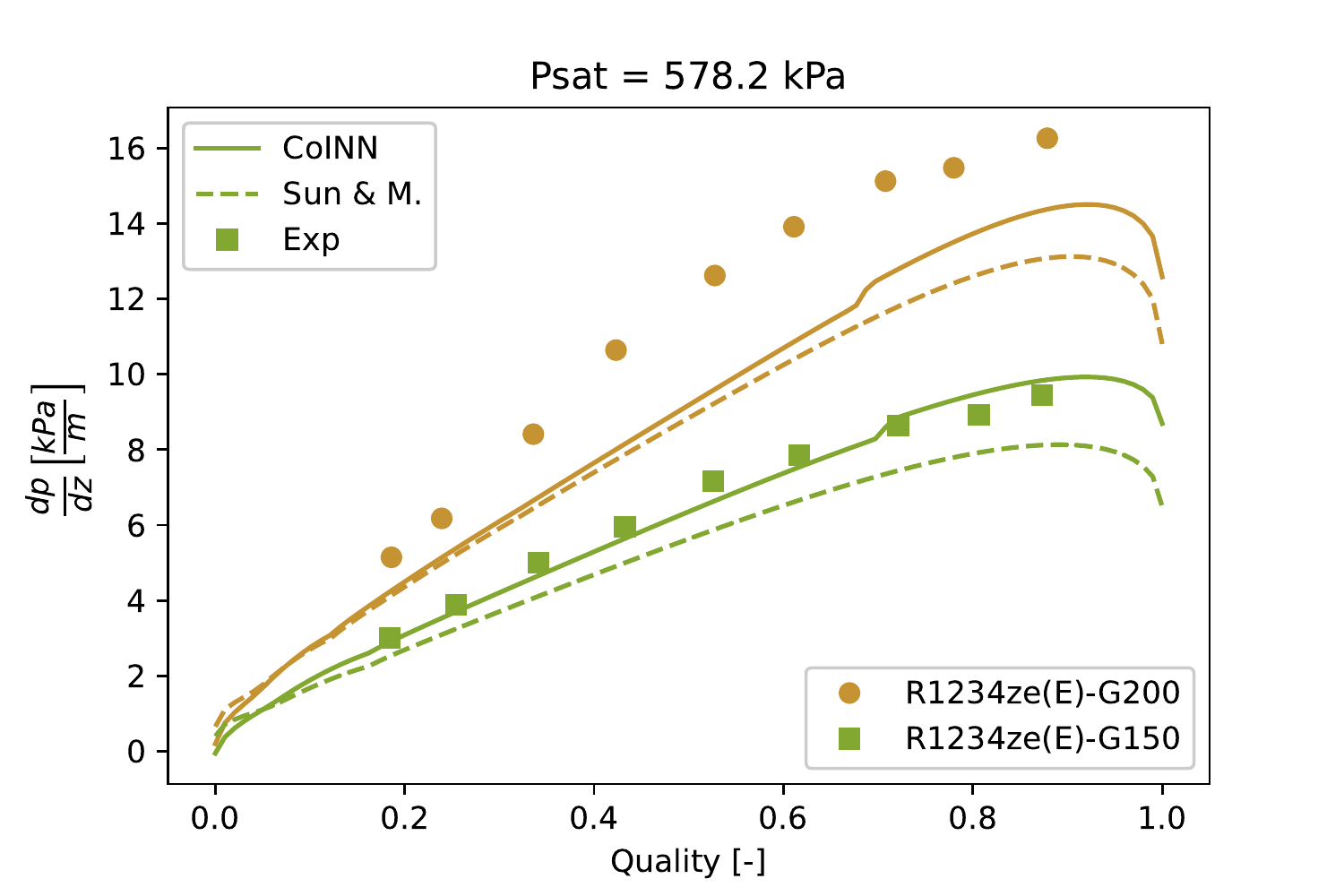}
\caption{\label{Fig10}Comparison of the results of CoINN, the Sun \& Mishima correlation, and the experimental results of the pressure drop for R1234ze(E). The solid lines represent the ANN, the dashed lines the Sun \& Mishima correlation, and circles experimental results for a mass flux of 200 kg/s-m$^2$ and squares a mass flux of 150 kg/s-m$^2$.}
\end{figure}

Fig. \ref{Fig10} shows the predicted pressure drop for the working fluid R-1234ze(E) with a saturation pressure of 578.2 kPa, which is in the range of pressures used for training. Here, for the mass flux of 150 kg/s-m$^2$, the $mre$ is 5.0\% for the ANN and 15.4\% for the Sun \& Mishima correlation. For the case of the mass flux of 200 kg/s-m$^2$, the $mre$ for the ANN is 18.5\% while for the Sun \& Mishima correlation, 22.7\%. 

\begin{figure}[h]
\centering
\includegraphics[width=12cm]{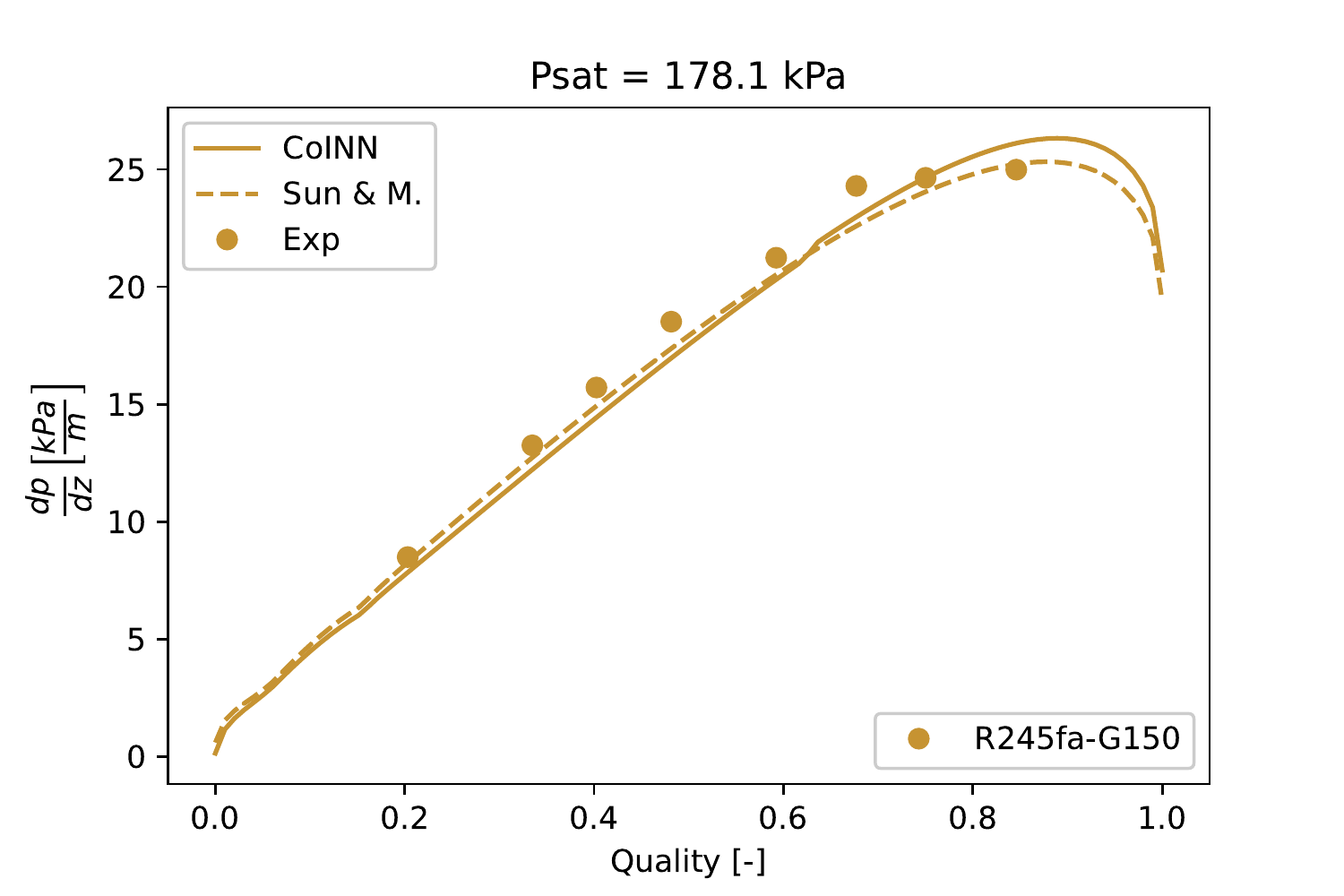}
\caption{\label{Fig11}Comparison of the ANN (solid line), the Sun \& Mishima correlation (dashed line), and experimental results (circles) for the pressure drop of R245fa with mass flux of 150 kg/s-m$^2$ and a saturation pressure of 178.1 kPa.}
\end{figure}

Finally, in Fig. \ref{Fig11}, it is shown the results for working fluid R-245fa with mass flux 150 kg/s-m$^2$ and saturation pressure 178.1 kPa. In this case, the saturation pressure is out of the training range, and the working fluid is not used for training. Here, the $mre$ for CoINN is 5.8\% and for the Sun \& Mishima correlation 3.9\%. In this case, both models show good agreement with the experiment with slightly better performance for the Sun \& Mishima correlation. 

\section{Conclusions}

Modeling physical phenomena of fluid dynamics applications is a current mechanical engineering challenge. Accurate prediction of flow parameters such as the pressure drop or the heat transfer coefficient is still challenging due to the complexity of the governing equations involved. Still, two methodologies are used for this: physical modeling and phenomenological modeling. On the one hand, physical modeling allows the generalization of multiple configurations while empirical correlations do not so. The continuous investigation and proposal of novel correlations show a clear trend in improving the generality of its application. For example, the correlations shown in this paper allow different configurations in the tube diameter, the operating pressure, and even the working fluid, etc. Still, they generally lacks accuracy in particular applications. On the other hand, as mapping ANNs, phenomenological modeling allows the representation of specific problems with high accuracy, but commonly with a reduction in generalization. 

In this paper, we successfully proved the application of a novel approach called correlated-informed neural networks (CoINN) for mapping purposes. We applied it in the pressure drop prediction during forced boiling in micro-channels for different mixtures and experimental configurations. Here, the knowledge acquired through experience and experiments and encapsulated in thermodynamic correlations is transferred to an ANN model. Indeed, the proposed model combines the physical modeling by the pressure drop correlation and the phenomenological modeling by the experimental data used to train the ANN. Furthermore, our model simplified considerably the number of inputs and the number of neurons in the hidden layer to produce good experimental fitting. The final configuration of CoINN had three inputs the quality, the inner diameter, the Sun \& Mishima correlation, and one hidden layer with 6 hidden neurons.

The results of our model outperformed those for one of the best correlations used for pressure drop prediction in micro-channels, the Sun \& Mishima correlation. The $mre$ for the data used by training and the posterior test averages 6\% in CoINN while the Sun \& Mishima correlation averages 13\%. We tested the overfitting of our model with 5 independent tests on the three different experimental configurations and in all of them, the ANN gave better results than the correlation.

Finally, we evaluated the ANN power of generalization and accuracy using available experimental data of condensation of R-134a, R-245fa, and R-1234ze(E) and with a range of saturation pressure between 178.1 and 1317.9 kPa. From these properties, two of the working fluids were not used for training the ANN and the saturation pressure covers a broader range than the pressure used for training. From these experiments, the average $mre$ for the ANN is 12.5\%, while for the Sun \& Mishima correlation is 14.2\%. Therefore, transferring information from the correlation to the ANN allows generalization to new situations.

\section*{Code Availability}
The source code of the ANN to predict the pressure-drop in micro-channels, including the results of Keniar et al. and Barraza et al., can be found on GitHub: \url{https://github.com/alejomonbar/CoINN}

\section*{Acknowledgments}

J.A.M.- B. thanks the National Council of Science and Technology (CONACyT), Mexico, for his Assistantship No. CVU- 736083. We thank Universidad de Guanajuato for the support in the realization of this research. We thank CETYS Universidad for the support in the completion of this research.

Adrián Mota-Babiloni acknowledges grant Juan de la Cierva - Incorporación 2019 (IJC2019-038997-I) funded by the Spanish Research Agency (MCIN/AEI/10.13039/501100011033).

\section*{Declarations of interest}

Declarations of interest: none.

\section*{Funding}

This research did not receive any specific grant from funding agencies in the public, commercial, or not-for-profit sectors.

\bibliography{mybibfile}
\end{document}